\documentclass{article}

\usepackage[preprint,nonatbib]{neurips_2023}

\usepackage[utf8]{inputenc} 
\usepackage[T1]{fontenc}    
\usepackage{hyperref}       
\usepackage{url}            
\usepackage{booktabs}       
\usepackage{amsfonts}       
\usepackage{nicefrac}       
\usepackage{microtype}      
\usepackage{xcolor}         
\usepackage[backend=biber, style=numeric, sorting=none, maxnames=10]{biblatex}

\usepackage{tikz}
\usetikzlibrary{shapes,arrows,positioning}
\usepackage{subcaption}
\title{Federated Learning of Medical Concepts Embedding using BEHRT}

\author{%
  Ofir Ben Shoham \\
  Department of Software and Information Systems Engineering\\
  Ben-Gurion University of the Negev\\
  \texttt{benshoho@post.bgu.ac.il} \\
  \And
  Nadav Rappoport \\
  Department of Software and Information Systems Engineering\\
  Ben-Gurion University of the Negev\\
\texttt{nadavrap@bgu.ac.il} \\
}
\begin{document}

\maketitle

\begin{abstract}
Electronic Health Records (EHR) data contains medical records such as diagnoses, medications, procedures, and treatments of patients. This data is often considered sensitive medical information. Therefore, the EHR data from the medical centers often cannot be shared, making it difficult to create prediction models using multi-center EHR data, which is essential for such models' robustness and generalizability. Federated Learning (FL) is an algorithmic approach that allows learning a shared model using data in multiple locations without the need to store all data in a central place.
An example of a prediction model's task is to predict future diseases. More specifically, the model needs to predict patient's next visit diagnoses, based on current and previous clinical data. Such a prediction model can support care providers in making clinical decisions and even provide preventive treatment. We propose a federated learning approach for learning medical concepts embedding. This pre-trained model can be used for fine-tuning for specific downstream tasks. Our approach is based on an embedding model like BEHRT, a deep neural sequence transduction model for EHR. We train using federated learning, both the Masked Language Modeling (MLM) and the next visit downstream model. We demonstrate our approach on the MIMIC-IV dataset. We compare the performance of a model trained with FL against a model trained on centralized data. We find that our federated learning approach reaches very close to the performance of a centralized model, and it outperforms local models in terms of average precision. We also show that pre-trained MLM improves the model’s average precision performance in the next visit prediction task, compared to an MLM model without pre-training. Our code is available at \href{https://github.com/nadavlab/FederatedBEHRT}{https://github.com/nadavlab/FederatedBEHRT}.
\end{abstract}

\section{Introduction}

Electronic Medical Records (EMR) or Electronic Health Records (EHR) is a collection of pieces of information documenting a patient's medical history (for example, a patient's visits and hospitalizations in a hospital). The medical records stored in hospitals contain critical medical information about the treatment protocol and its results \cite{evans2016electronic}.

Multi-center studies have the potential to enhance models' ability to capture and adapt to heterogeneity, leading to an improvement in their generalizability. Furthermore, collecting data from multiple sources results in a larger dataset for training prediction models, which reduces the expected generalization error and increases the robustness of the model \cite{dang2022federated}. In addition, rare conditions may not be represented well enough in a single dataset, but using data from multiple sources may increase the statistical power \cite{pati2022federated}.

EHRs contain sensitive medical information, which can make it challenging to share among healthcare providers \cite{bani2020privacy}. Federated learning \cite{mcmahan2017communication} is an algorithmic approach that trains a single model based on several databases stored in separate locations (clients) without consolidating the information in one central location. This approach makes it possible to train a shared global machine learning model with the help of a central server without sharing the observations between the different databases. In particular, federated learning is suitable for training a computational model based on information sources from separate medical centers (multi-center study) while maintaining the privacy of data, patients, and medical centers \cite{xu2021federated}.

An example of a prediction task based on EHR data is the prediction of future diagnoses, also called next visit prediction. In this task, we want to train a model that can predict the diagnoses of a patient that will be diagnosed in their next visit based on current and previous clinical data.

BEHRT \cite{peng2020behr} is a deep neural sequence transduction model based on BERT \cite{devlin2018bert} model architecture for EHR. The input for this model is a sequence constructed with words representing diagnoses, sentences representing each visit, and a document representing a patient's complete medical history. In their work, they first trained a Masked Language Modeling (MLM) model and then used it as a pre-trained model and fine-tuned it for the next visit prediction task. Afterward, They demonstrated their approach on the CPRD dataset that contains medical records from general practitioners \cite{herrett2015data}. BEHRT demonstrates an enhancement of 8.0--13.2\% (in terms of average precision scores for various tasks) compared to the state-of-the-art deep EHR models like RETAIN \cite{choi2016retain} and Deepr \cite{miotto2016deep} models \cite{peng2020behr}. The BEHRT architecture is designed to easily incorporate multiple heterogeneous medical concepts, including diagnoses, measurements, and more. Another advantage of BEHRT is that it results in an interpretable model \cite{vig2019bertviz}, which is crucial for clinicians to understand why the model arrives at its predictions. Furthermore, BEHRT's patient representation can be used as a pre-trained model for downstream tasks \cite{peng2020behr}. 


We propose federated learning training to medical concepts embedding using BEHRT. Our approach utilizes federated learning training to enhance the robustness and generalizability of the BEHRT model. The architecture of the BEHRT model, as discussed in \cite{dang2022federated}, is limited by centralizing all data, which prevents the BEHRT model from handling multi-center data. Our approach used federated learning training for the pre-trained MLM phase and also for the next visit prediction task. Our approach is applicable to any dataset containing clinical data per patient and is suitable for multi-center studies that require a federated learning algorithm to ensure EHR data privacy.

We demonstrated our approach using the MIMIC-IV dataset \cite{johnson2020mimic} for the next visit prediction task. Our federated learning approach improves average precision by 4-10 absolute percents compared to local models, and achieves very close average precision performance to centralized models, while maintaining data privacy and scalability for multi-center studies.

\section{Related work}
Our work relies on a representation model of medical concepts. In recent decades, word2vec methods \cite{mikolov2013efficient} have gained popularity not only in classical NLP but also in Precision Medicine \cite{si2021deep}. For example, Phe2vec \cite{douglas2020phe2vec} created patient embeddings by representing the patient's medical concepts according to intervals of days. Each interval is represented by a sentence, and each word in the sentence is one medical concept. After representing the medical information as a sequence, different word2vec \cite{mikolov2013efficient} methods can be used, such as Glove \cite{pennington2014glove} and FastText \cite{busta2015fastext} and also BERT \cite{devlin2018bert} based on transformers \cite{vaswani2017attention}. BRLTM utilized transformers in clinical prediction models \cite{wu2020bidirectional} to predict depression. They trained a transformer model with MLM and afterward fine-tuned the pre-trained model using the depression diagnoses task. However, they used LDA \cite{blei2003latent} for clinical notes representation, which is bag-of-words representation. Instead of LDA, HORDE model \cite{lee2020harmonized} used LSTM \cite{hochreiter1997long} for clinical notes, which is not bag-of-words. However, there are better alternatives for text representation such as ClinicalBert \cite{huang2019clinicalbert} based on BERT \cite{devlin2018bert} which outperforms bag-of-words representations, as well as Bi-LSTM language models.

Another common transformer-based model, is Med-BERT \cite{rasmy2021med}. They trained BERT \cite{devlin2018bert} model using the MLM task, then trained the model for length-of-stay task and demonstrated an improvement on two downstream tasks compared to GRU \cite{chung2014empirical} and RETAIN \cite{choi2016retain}, but there is no comparison to BEHRT. The main differences between Med-BERT and BEHRT \cite{peng2020behr} is that Med-BERT was trained also on the length-of-stay task and has more training samples compared to BEHRT. However, Med-BERT has a ranking for each event, and the ranking of the importance of each event has not been studied enough \cite{rasmy2021med}. In addition, They did not include the time between different visits, unlike BEHRT. Therefore we chose to illustrate our approach using BEHRT.

In addition to text-based methods, there are also methods that model EHR data using graph representation \cite{zhang2021harmonized,wanyan2021deep, wanyan2022relational, pellegrini2022unsupervised} and also graph-based methods that used a Knowledge Graph \cite{chen2020knowledge} such as \cite{nelson2022embedding, baranzini2022biomedical}. However, a fundamental disadvantage of using a Knowledge Graph is that it required to validate the graphical model \cite{baranzini2022biomedical}. In addition, such graph-based methods may have memory limitations, as the entire graph containing all patients often cannot fit into memory \cite{pellegrini2022unsupervised, wanyan2022relational}, taking up a considerable amount of both time and running memory \cite{duan2022comprehensive}.

Med-BERT \cite{rasmy2021med} used multi-center data, but all the data located in one central location, which limits the available data due to concerns over infrastructures, regulations, privacy, and data standardization present a challenge to data sharing across healthcare institutions \cite{dang2022federated}. Multi-center EHR data enables the larger and varied data for model training which is essential in order to improve model generalizability and robustness \cite{dang2022federated}. There are federated algorithms to overcome this limitation such as \cite{mcmahan2017communication, li2020federated, boughorbel2019federated, hsu2019measuring}. Dang et al. \cite{dang2022federated} compared multiple federated learning algorithms such as FedAvg, FedAvgM, FedProx, FedAdam and FedAdagrad. Among the federated learning algorithms, the FedAvg and FedAvgM algorithms achieved slightly better results than FedProx, FedAdam and FedAdagrad \cite{dang2022federated}. Therefore, In our work, we use FedAvg federated learning algorithm for the MLM and next visit prediction tasks. We used transformer-based modeling according to the model architecture of BEHRT \cite{peng2020behr}.

\section{Methods}
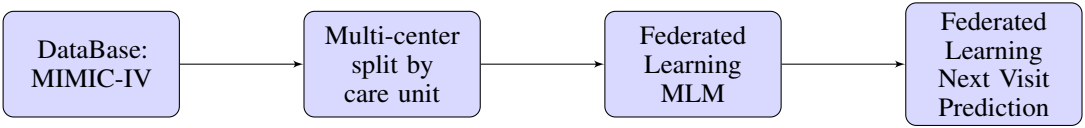
\begin{figure}[htbp]
    \centering
    \begin{tikzpicture}[node distance=4cm]
        \tikzstyle{block} = [rectangle, draw, fill=blue!15, 
            text width=6em, text centered, rounded corners, minimum height=4em]
        \tikzstyle{line} = [draw, -latex']
        
        \node [block] (db) {DataBase: MIMIC-IV};
        \node [block, right of=db] (center) {Multi-center split by care unit};
        \node [block, right of=center] (fl) {Federated Learning MLM};
        \node [block, right of=fl] (bert) {Federated Learning Next Visit Prediction};
        
        \path [line] (db) -- (center);
        \path [line] (center) -- (fl);
        \path [line] (fl) -- (bert);
    \end{tikzpicture}
    \caption{stages in our study}
    \label{fig:research-steps}
\end{figure}

An overview of the stages of this study is illustrated in  Figure~\ref{fig:research-steps}. Initially, we retrieved the data from the raw source MIMIC-IV database \ref{subsec:Data}. Next, we simulate a federated data scenario by dividing the data into multiple centers. Each patient was assigned to a single center according to the center where it had the longest stay. Afterward, we employed federated learning training for Masked Language Modeling (MLM). Lastly, we utilized the MLM pre-trained model for federated learning of the next visit prediction task.

\subsection{Next Visit problem definition}
Let $P$ denote the set of patients and let each patient $p$ have medical data consisting of $n$ visits: $V_{p} = {V_{1,p}, V_{2,p}, \ldots, V_{n,p}}$. For a given visit $i$ of patient $p$, $V_{i,p}$ represents the set of diagnoses assigned to patient $p$ at visit $i$. Specifically, $V_{i,p} = {d_{1,i,p}, d_{2,i,p}, \ldots, d_{m,i,p}}$, where $m$ is the number of diagnoses assigned to patient $p$ at visit number $i$. In the next visit prediction task, we choose a random $j$ visit number, assuming we have the medical data until ${V_{j}}$, we need to predict the diagnoses ${{d_{1,j+1,p}}, {d_{2,j+1,p}}, \ldots, {d_{m,j+1,p}}}$ for visit number ${j+1}$ based on ${V_{1,p}, V_{2,p}, \ldots, V_{j,p}}$.

\subsection{Data}
\label{subsec:Data}
MIMIC-IV is a comprehensive healthcare dataset that was utilized to demonstrate the usability of our suggested approach. MIMIC-IV contains about 400,000 ICU admissions of about 190K patients from Beth Israel Deaconess Medical Center in Boston, Massachusetts spanning a period of five years from 2008 to 2012 \cite{johnson2020mimic}.
Figure~\ref{fig:mimic_iv_flowchart} provides an overview of the process of identifying patients with ICD10 diagnoses in the MIMIC-IV database and the steps involved in aggregating the data to simplify the representation of the diagnoses. At the top of Figure~\ref{fig:mimic_iv_flowchart}, we have 190,279 patients with admissions. From this group, 84,453 were found to have an ICD10 code associated with their condition, while 105,826 did not have an ICD10 diagnoses code. These patients had 17,009 different ICD10 diagnosis codes. These codes were then aggregated into 416 groups according to the Clinical Classifications Software (CCS) of Healthcare Cost and Utilization Project (HCUP). Finally, each observation is a sequence that represents the medical history of a single patient, which includes his diagnoses, age, and year of diagnosis. The data for each patient is actually composed of multiple visits, ordered by admission start time, which is important for the next visit prediction task.

\tikzstyle{startstop} = [rectangle, rounded corners, minimum width=3cm, minimum height=1cm, text centered, draw=black, fill=red!30]
\tikzstyle{process} = [rectangle, minimum width=3cm, minimum height=1cm, text centered, draw=black, fill=orange!30]
\tikzstyle{arrow} = [thick,->,>=stealth]

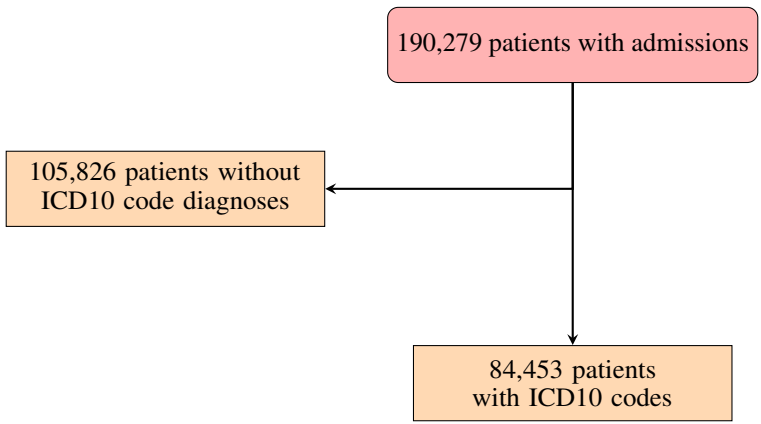
\begin{figure}[ht!]
\centering
\begin{tikzpicture}[node distance=2cm]

\node (admissions) [startstop] {190,279 patients with admissions};
\node (no-icd10) [process, below left of=admissions, xshift=-4cm, yshift=-0.5cm, text width=4cm, align=center] {105,826 patients without ICD10 code diagnoses};
\node (icd10) [process, below of=admissions, yshift=-2.5cm, text width=4cm, text width=4cm, align=center] {84,453 patients with ICD10 codes};

\draw [arrow] (admissions) |- (no-icd10);
\draw [arrow] (admissions) -> (icd10);

\end{tikzpicture}
\caption{Patient admissions and diagnoses flowchart.}
\label{fig:mimic_iv_flowchart}
\end{figure}

\subsubsection{Multi-Centers Split}
To demonstrate the need for federated learning for the next visit prediction task, we simulated a multi-centers scenario by splitting our data by patient. To simulate a real-world biased variety between medical centers we did not split the patients randomly but clinically-driven. Each patient was assigned to a single care unit according to the unit with the longest stay. Length of stay was taken from the MIMIC-IV transfers table in Hosp module \cite{johnson2020mimic}. After splitting the patients into centers, we obtained a total of 39 centers.

\subsection{Baseline Approaches}
In order to compared our federated learning approach we trained a centralized model. In the centralized training, the two learning phases of MLM and next visit prediction were trained using a single dataset covering all the training samples.
In addition, we also compared our approach to local model training. In the local training, no information is shared across clients. As we have 39 centers, we trained each center's model separately using its local data - first we trained MLM for the local data, and then we fine-tuned the MLM model using the client's local data for the next visit prediction. 

\subsection{BEHRT}
We used the BEHRT \cite{peng2020behr} model architecture for federated learning for both the MLM and the next visit prediction downstream task. BEHRT is a deep learning language model based on the BERT architecture \cite{devlin2018bert}. BEHRT consists of Masked Language Modeling (MLM), followed by fine-tuning the pre-trained MLM model for the next visit prediction task. In the MLM training, the task is to predict the masked disease tokens. The features for the MLM tasks are: diagnoses, patient's age, and the diagnosis year. For the next visit task the features are the same as those for the MLM, but the list of diagnoses is partial and contains the medical information up to the visit for which we want to predict its diagnoses. In the MLM phase, the model learns an embedding of the clinical concepts such as diagnosis, age, position (i.e., the relative position of a concept within a visit), and segment (i.e., visit). Afterward, the MLM is fine-tuned for the next visit prediction task by adding a classification layer \cite{peng2020behr}.

\subsection{Our Approach}
\label{subsec:our_approach}

\begin{figure}[!ht]
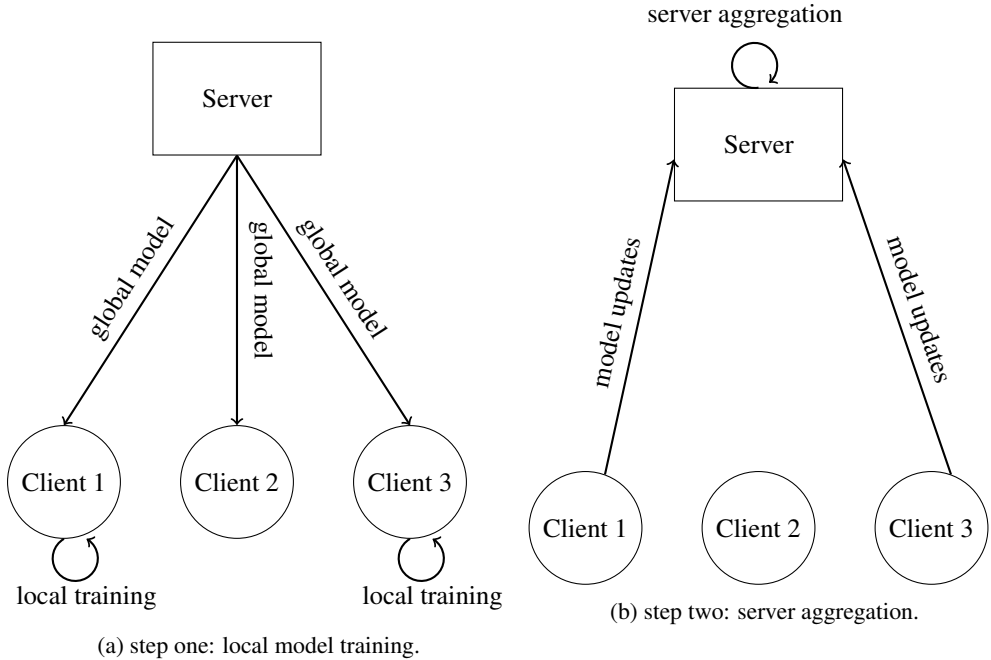

    \centering
    \begin{subfigure}{0.5\textwidth}
    \centering
    \begin{tikzpicture}[node distance=3.5cm]
    \node[rectangle, draw=black, text width=2cm, minimum height=1.5cm, align=center] (server) {Server};
    
    \node[circle, draw=black, minimum size=1.5cm, below of=server, xshift=-2.3cm, yshift=-1.6cm] (client1) {Client 1};
    
    \node[circle, draw=black, minimum size=1.5cm, below of=server, xshift=0cm, yshift=-1.6cm] (client2) {Client 2};
    \node[circle, draw=black, minimum size=1.5cm, below of=server, xshift=2.3cm, yshift=-1.6cm] (client3) {Client 3};
    
    \draw[->, thick] (server.south) -- (client1.north) node[midway, above, sloped] {global model};
    \draw[->, thick] (server.south) -- (client2.north) node[midway, above=0.03, sloped] {global model};
    \draw[->, thick] (server.south) -- (client3.north) node[midway, above, sloped] {global model};
    \draw[->, thick] (client1.south) arc (-240:60:0.3) node[below=0.5] {local training};
    \draw[->, thick] (client3.south) arc (-240:60:0.3) node[below=0.5] {local training};

    \end{tikzpicture}
    \caption{step one: local model training.}

    \label{fig:server_sends_global_model_client_local_training}
    \end{subfigure}
    \begin{subfigure}{0.45\textwidth}
    \begin{tikzpicture}[node distance=3.5cm]
    \node[rectangle, draw=black, text width=2cm, minimum height=1.5cm, align=center] (server) {Server};
    
    \node[circle, draw=black, minimum size=1.5cm, below of=server, xshift=-2.3cm, yshift=-1.6cm] (client1) {Client 1};
    
    \node[circle, draw=black, minimum size=1.5cm, below of=server, xshift=0cm, yshift=-1.6cm] (client2) {Client 2};
    \node[circle, draw=black, minimum size=1.5cm, below of=server, xshift=2.3cm, yshift=-1.6cm] (client3) {Client 3};

    \draw[->, thick] (client1.70) -- (server.-170) node[midway, above, sloped] {model updates};
    \draw[->, thick] (client3.70) -- (server.-10) node[midway, above, sloped] {model updates};
    
    \draw[->, thick] (server.north) arc (280:-50:0.3) node[midway, right=0.2cm, above=0.1] {server aggregation};
    
    \end{tikzpicture}
    \caption{step two: server aggregation.}
    \vphantom{\includegraphics[height=0.30cm]{example-image-a}}
    
    \label{fig:clients_send_updates_server_aggergates}

    \vphantom{\includegraphics[height=0.7cm]{example-image-a}}

    \end{subfigure}
\caption{Federated Learning algorithm for MLM and next visit prediction. (a) In the first step, the server sends the global model to all clients, and each selected client trains the local model. (b) In the second step, the server gets the trained weights from the selected clients, aggregates the weights and updates the global model.}
\label{fig:clients_train_on_local_data}

\end{figure}

We used the Federated Averaging (FedAvg) algorithm \cite{mcmahan2017communication} for BEHRT federated learning. The server initially shares the BEHRT global model with each client. Subsequently, the selected client trains the model using their local data as depicted in Figure~\ref{fig:server_sends_global_model_client_local_training}. In the second part of the algorithm, the selected clients transmit the weights of their trained local model to the server, and the server updates the global model by aggregating all the updated models by computing a weighted average of each weight according to client's sample size, as shown in Figure~\ref{fig:clients_send_updates_server_aggergates}. Finally, the server disseminates the updated model to all the clients. This process continues iteratively until a stop criterion is met. We used this federated learning algorithm for both the MLM training step and the next visit prediction model training step. At each round of training, we selected only a fraction of 10\% from the clients to train on their local data. We did this for efficiency, as \cite{mcmahan2017communication} showed that there is a point of diminishing returns when adding more clients.

\section{Experiments}

\subsection{Experimental Setup}
We performed experiments to compare our proposed federated learning approach to a model trained with all the data in a central place. To assess the performance of our proposed approach, we partitioned the data into train and test sets using an 80-20 split ratio. We trained the MLM model using PyTorch, based on the train set with Adam optimizer \cite{kingma2014adam} and 500 epochs. We selected the best model based on the precision on the validation set. Training the MLM model with the centralized data took about three hours to reach the best model on an RTX 3090 GPU. The FL MLM training took about 14 hours. The runtime of the next visit prediction (FL and Non-FL) took between a few minutes and up to 2 days, depending on the exact configuration (see Supplementary). The FL training takes longer than centralized training because during each round of training, we sequentially train a subset of clients on a single GPU. As discussed in section \ref{subsec:our_approach}, during each round of training, we randomly select a subset of clients and train the model on their local data. Consequently, the FL training procedure takes more time compared to traditional centralized learning approaches. We repeated the training of the next visit prediction while varying the random seed in order to calculate the confidence intervals.

\begin{figure}[ht]
    \centering
    \includegraphics[width=1\textwidth]{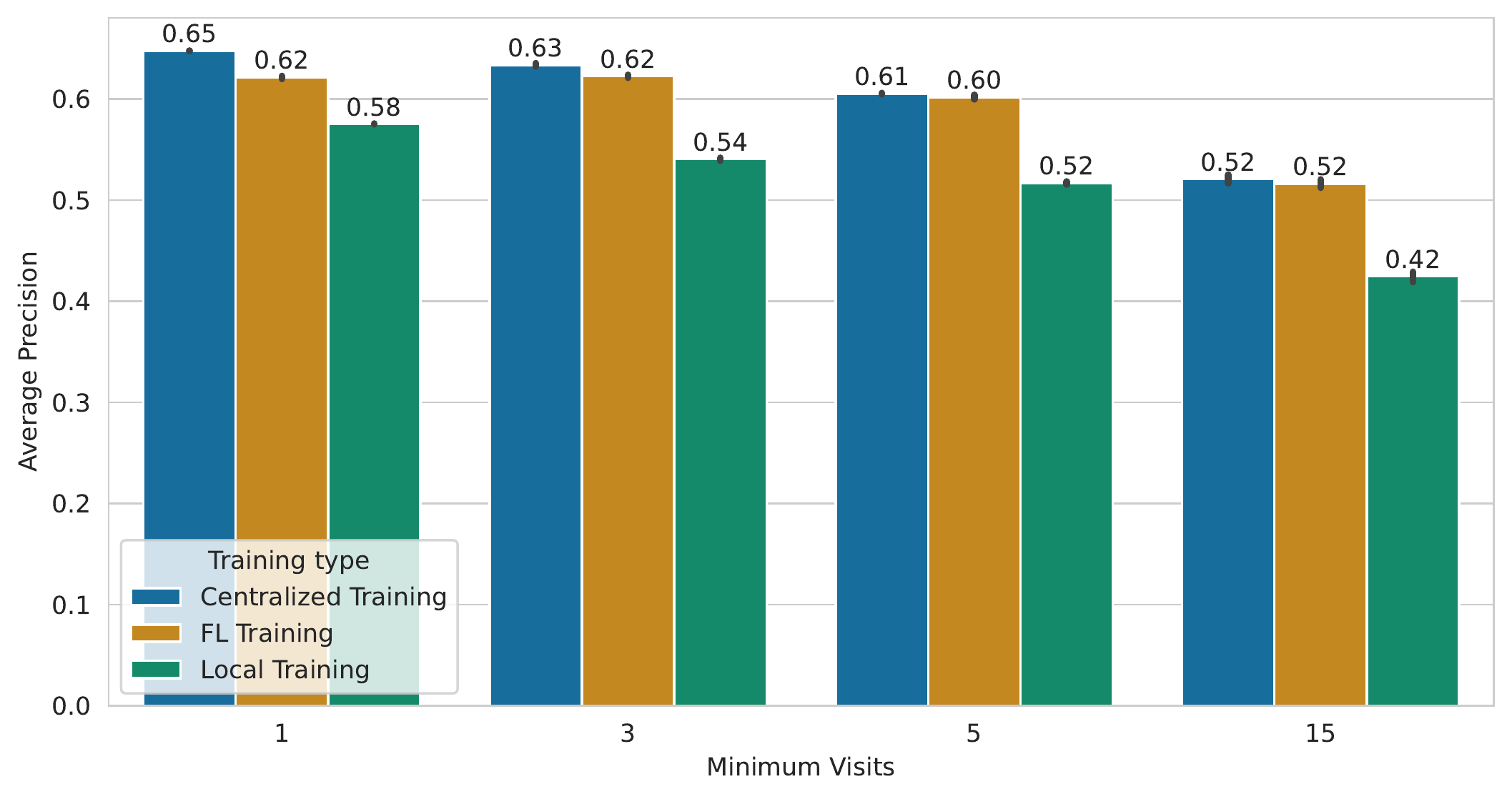}
    \caption{The average precision of each training method was evaluated for the next visit prediction task. The centralized model is referred to as Centralized Training, our proposed approach is FL Training, and Local Training involves training local models in our multi-center study. We evaluated the average precision of the models at four minimum visit thresholds. The average precision value appears at the top of each bar plot, and also the 95\% confidence interval based on a random seed.}
    \label{fig:ap_fl_vs_non_fl_vs_local_models_barplot}
\end{figure}

\subsection{Federated vs. centralized Learning}
In this experiment, we compared our proposed approach (FL Training) to a model trained with centralized data. For the FL training, the two phases were trained using the federated data. We trained a single MLM model and multiple next-visit models where in each model, we subset the data to patients having at least 1, 3, 5, or 15 visits. Our results showed that our proposed FL model achieved similar average precision to the centralized model for minimum visits of 3, 5, and 15. For a minimum visit of 1, the centralized model outperformed our model by absolute 3\% (Figure~\ref{fig:ap_fl_vs_non_fl_vs_local_models_barplot}).

\subsection{Federated vs. local client-independent Learning}
This experiment simulates a scenario where no data can be shared due to privacy and security concerns, making local model training a common scenario in such cases. Each local model was trained with its own local data, which varied in size and clinical conditions. To aggregate the performances of the local models, we used weighted averages based on their average precision and the number of examples (patients) in the local train data.
Figure~\ref{fig:ap_fl_vs_non_fl_vs_local_models_barplot} shows the average precision results of local training compared to FL training and centralized training for four minimum visit thresholds. Our federated learning approach outperformed local training for minimum visits of 1, 3, 5, and 15 by an  average precision of absolute 4\%, 8\%, 8\%, and 10\%, respectively. Overall, our proposed FL training model achieved 4-10\% absolute higher average precision than local training models.

\begin{figure}[ht]
    \centering
    \includegraphics[width=1\textwidth]{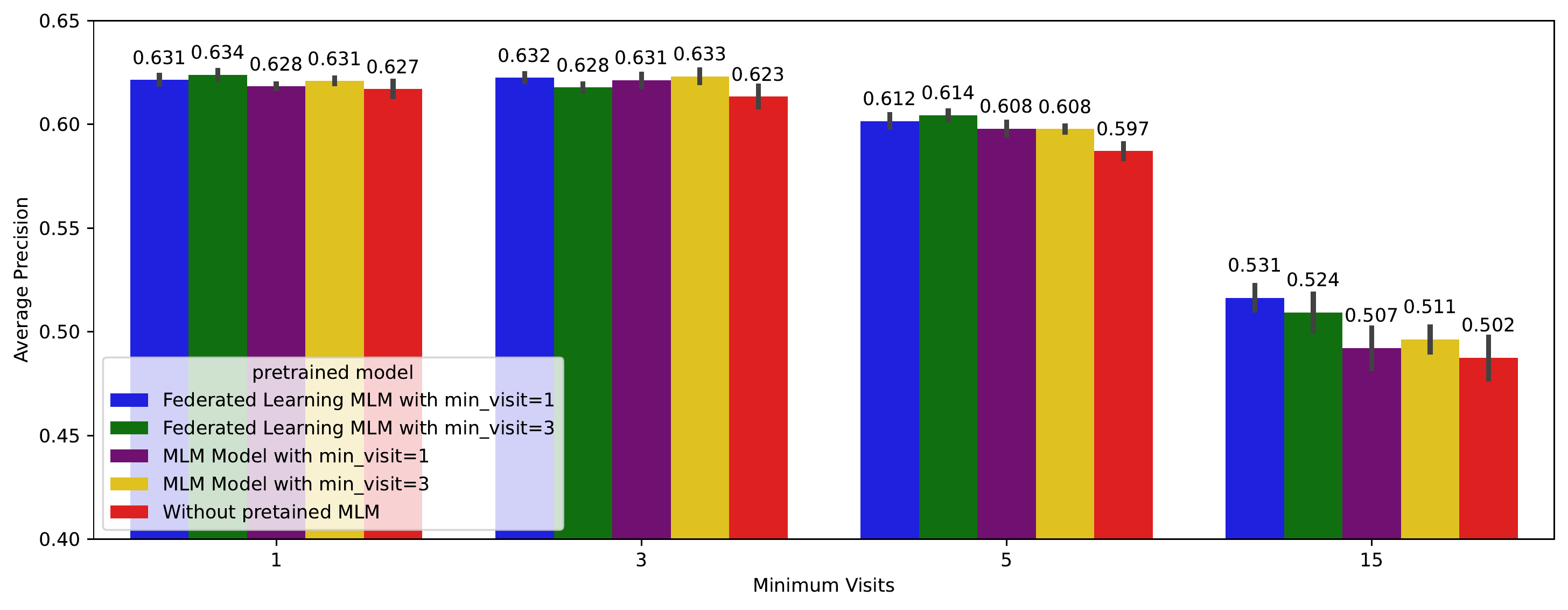}
    \caption{The importance of the pre-trained MLM for the FL next visit prediction task. We compared the performance for five MLM configurations: blue and green with pre-trained MLMs and fine-tuned with federated learning (for patients with minimum visits of one or three, respectively); purple and yellow with centralized MLM training (also for patients with minimum visits of one or three, respectively); and red without pre-trained MLMs.}
    \label{fig:pretained_mlm_importance}
\end{figure}

\subsection{Pre-trained MLM}
In the next step, we took the pre-trained MLMs and fine-tuned them for the prediction task. In this experiment, we conducted an ablation study to evaluate the importance of pre-trained MLM. Specifically, we compared the performance of FL next visit prediction using different pre-trained MLM models. We evaluated two centralized MLMs: the first was an MLM model with a minimum of one visit (trained on all patients), and the second was an MLM model trained on patients who had at least three visits. Additionally, we evaluated two more FL MLM models. The first FL MLM is the one trained with patients with a minimum of one visit, and the second is for patients with at least three visits. Finally, we compared the performance of all these pre-trained models to the performance of the model without pre-trained MLM. Figure~\ref{fig:pretained_mlm_importance} shows the average precision comparison of FL next visit prediction based on the pre-trained MLM models. This figure shows that for minimum visits of 3 and 5, the pre-trained MLM improves the average precision for FL next visit prediction by 1-1.2\% absolute compared to without pre-trained MLM. Moreover, the difference in average precision between the centralized MLM and FL MLM was negligible. These findings indicate that FL MLM can achieve similar performance without having all the data centralized in one place.




\section{Discussion}
In this paper, we present a federated learning approach for BEHRT. We trained the MLM and the next visit prediction task using the FedAvg algorithm \cite{mcmahan2017communication}. Our approach is general and well-suited for multi-center studies that require a federated learning model to ensure the privacy of EHR data. We show that our approach of federated learning of embedding clinical concepts can meat the performance of a model trained on centralized data, and it outperforms model trained locally with no information sharing. We demonstrate the effectiveness of our approach by simulating the MIMIC-IV dataset as a multi-center study, training a federated learning MLM and next visit prediction models. We compare the performance of our federated learning approach to both a centralized model and local models (which are commonly used due to data privacy concerns).

In the first experiment, we compare the average precision of FL training, centralized training, and local training for different minimum visit thresholds. As previously mentioned, our federated learning approach achieved average precision results that were comparable to the centralized baseline approach. For minimum visit thresholds of 3, 5, and 15, the differences in average precision were negligible. These results demonstrate that our approach can achieve similar performance to centralized training while preserving EHR data privacy. The reason for lower performance for a minimum visit threshold of 1 is not clear enough. One possible reason is that the set of diagnoses of the patients in this dataset are more diverse, which could make it more difficult for the FL model to generalize well across all clients. In contrast, for minimum visits threshold of 3 and above the sample size is smaller and the set of possible diagnoses and concept to learn their embedding is smaller. 

We compare our approach to local models, where each center trains with its local data. We find that the difference in performance between local training and our federated learning approach increased as the minimum visit threshold increased from 1 to 3 and from 5 to 15 (Figure \ref{fig:ap_fl_vs_non_fl_vs_local_models_barplot}). A possible reason for the decrease in performance of the local models when increasing the threshold of minimum visits, is because local models have less data, making it challenging to learn a local model with good performance. In contrast, the difference between our federated learning approach and local models is much more significant when there is less data in each center, because the federated learning approach deals with this by learning a common model, while the local model will have less robustness when it has few examples. In addition, the average precision of the next visit models is lower when the minimum number of visits increases. We believe this is because the number of samples decreases as the minimum visit threshold increases.

In our second experiment, we investigated the impact of using different fine-tuned masked language models (MLMs) for predicting the next medical visit with federated learning. We found that the performance of centralized MLMs and federated MLMs was similar, but both outperformed the models without pre-trained MLMs (Figure \ref{fig:pretained_mlm_importance}). These results demonstrate that pre-training the MLMs can significantly improve the average precision of the next visit prediction models. Furthermore, we observed that the performance gap between the pre-trained MLMs and the models without pre-training increased as the minimum number of visits per patient increased. This is may be because the pre-trained MLMs are particularly valuable in low-data scenarios, where the pre-trained MLMs can help to improve the generalization and robustness of the models. Moreover, it can be seen that FL MLM has better performance than a centralized MLM as a pre-trained MLM model for fine-tuning for the FL next visit prediction (comparing the blue and green bars to purple and yellow bars on Figure \ref{fig:pretained_mlm_importance}). 

One potential direction for future work is to apply the federated learning approach proposed in this study to real multi-center data, in order to assess its performance in a real-world scenario. Additionally, future work could consider combining a wider range of features, such as laboratory results and vital signs, to further improve the accuracy of the predictive models. Another interesting direction for future research would be to investigate the potential of applying alternative models, such as Med-BERT \cite{rasmy2021med}, to the federated learning approach presented in this study.

\section{Conclusion}
In this study, we proposed a federated learning approach using the FedAvg algorithm to train a masked language model and a next visit prediction task, enabling the privacy of EHR data to be maintained in multi-center studies. Our federated learning approach achieved similar performance to the centralized model, and an improvement of 4-10 absolute percents of average precision compared to local models. This highlights the importance of our federated learning approach for creating a common model for multi-center studies while preserving data privacy and improving the generalizability and robustness of the model. Furthermore, our approach is general to any multi-center study and it is scalable to any number of clients, compared to local models and the centralized model baseline approaches.


\printbibliography
\end{document}